\documentclass[a4paper,twoside]{article}

\usepackage[T1]{fontenc}
\usepackage[utf8]{inputenc}
\usepackage{graphicx}
\usepackage{xcolor} 
\usepackage{amsmath,amssymb,amsthm,amstext}
\usepackage{calc}
\usepackage{multirow}
\usepackage{subcaption}
\usepackage[bottom]{footmisc}
\usepackage{url}
\usepackage{float} 
\usepackage[ruled,vlined]{algorithm2e}

\makeatletter
\newcommand{\authorname}[1]{\begin{center}\large\textbf{#1}\end{center}}
\newcommand{\affiliation}[1]{\begin{center}\itshape #1\end{center}}
\newcommand{\email}[1]{\begin{center}\texttt{#1}\end{center}}
\renewcommand{\abstract}[1]{\small\noindent\textbf{Abstract:} #1\par\normalsize}
\newcommand{\keywords}[1]{\small\noindent\textbf{Keywords:} #1\par\medskip\normalsize}
\makeatother

\newcommand{\safeincludegraphics}[2][]{%
  \IfFileExists{#2}{\includegraphics[#1]{#2}}{%
    \fbox{\parbox{0.95\linewidth}{\ttfamily Missing file: #2}}%
  }%
}

\title{Data Augmentation Supporting a Conversational Agent Designed for Smoking Cessation Support Groups}

\begin{document}

\maketitle

\authorname{Salar Hashemitaheri and Ian Harris}
\affiliation{University of California, Irvine}
\email{salarh@uci.edu, harris@ics.uci.edu}


\abstract{
Online support groups for smoking cessation are economical and accessible, yet they often face challenges with low user engagement and stigma. The use of an automatic conversational agent would improve engagement by ensuring that all user comments receive a timely response. A required step for a conversation agent is intent classification which classifies the meaning of each participant's utterance to inform the agent about how to respond. A limitation in the development of intent classification models is the low availability of sufficient labeled data which can be used to train the model. To mitigate the effects of noisy and sparse data, we fine-tune a pretrained large language model (LLM). We address the challenge of insufficient high-quality data by employing a two-level data augmentation strategy: synthetic data augmentation and real data augmentation. First, we fine-tuned an open source LLM to classify posts from our existing smoking cessation support groups and identify intents with low F1 (precision+recall) scores. Then, for these intents, we generate additional synthetic data using prompt engineering with the GPT model, with an average of 87\% of the generated synthetic posts deemed high quality by human annotators. Overall, the synthetic augmentation process resulted in 43\% of the original posts being selected for augmentation, followed by 140\% synthetic expansion of these posts. Additionally, we scraped more than 10,000 real posts from a related online support context, of which 73\% were validated as good quality by human annotators. Each synthetic or scraped post underwent rigorous validation involving human reviewers to ensure quality and relevance. The validated new data, combined with the original support group posts, formed an augmented dataset used to retrain the intent classifier. Performance evaluation of the retrained model demonstrated a 32\% improvement in F1, confirming the effectiveness of our data augmentation approach. Synthetic and real post augmentation led to similar performance improvements. This study provides a replicable framework for enhancing conversational agent performance in domains where data scarcity is a critical issue.
}

\section{Introduction}

Smoking cessation remains one of the most critical public health challenges, driven by the profound health risks associated with tobacco use. However, the journey to quit smoking is often fraught with difficulties, not least of which is the pervasive stigma attached to being a smoker. This stigma, manifesting as societal judgment and negative stereotypes, exacerbates feelings of shame, guilt, and isolation among smokers. These emotions often deter individuals from seeking help or participating in cessation programs, further entrenching their habit and making quitting even more challenging (Lozano et al., 2020; Hosseinisangchi et al., 2024).

The stigma surrounding smoking is particularly damaging in social contexts where smokers may already feel marginalized. For instance, smokers who are trying to quit might hesitate to join support groups, fearing judgment from non-smokers or even their peers within these groups (Whittaker et al., 2018). This fear of being stigmatized can lead to reduced engagement in support activities, decreased participation in discussions, and an overall reluctance to seek help. Without active engagement, the effectiveness of these support groups is significantly diminished, and the opportunity to create a supportive community that can counteract stigma is lost (Kim et al., 2018).

In response to these challenges, online support groups have emerged as a vital resource for smokers seeking to quit. These platforms provide an economical and accessible means of offering both emotional and informational support. However, despite their potential, online support groups often suffer from low user engagement due to poor response times. Also, the stigma that online platforms aim to alleviate can still pervade these spaces, discouraging open discussion and participation (Rapp et al., 2021; Oveisi et al., 2023).

To address these issues, there is a growing interest in leveraging technology, specifically natural language processing (NLP) and large language models (LLMs), to enhance user interaction within these support groups (Bayer et al., 2021). One promising approach is the integration of NLP-based smart conversational agents into online support platforms (Whittaker et al., 2022). These conversational agents can serve as non-judgmental, always-available companions that engage users, detect their intents, and provide timely, evidence-based responses (Woebot (n.d.)). By offering personalized support tailored to the unique challenges faced by smokers, these agents can reduce stigma and promote user engagement.

Conversational agents, also known as chatbots or virtual assistants, are AI-powered systems designed to interact with users in a natural language format. These agents are increasingly used across various domains, including customer service, healthcare, and mental health support, due to their ability to provide immediate and personalized responses to user inquiries. A typical conversational agent processes user inputs, understands the intent behind the messages, and generates appropriate responses, mimicking human-like conversations.

\begin{figure}[htbp]
  \centering
  \safeincludegraphics[width=\columnwidth]{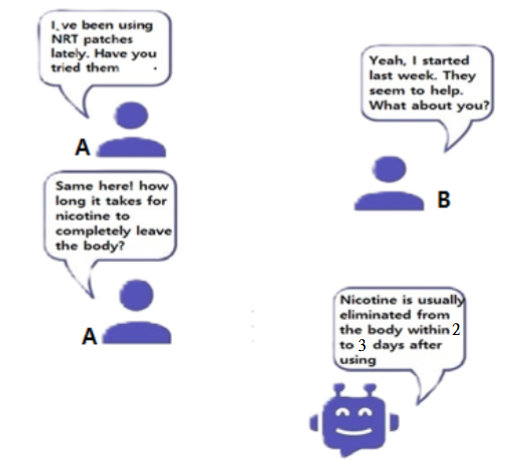}
  \caption{Conversational Agent}
  \label{fig:agent1}
\end{figure}

Figure~\ref{fig:agent1} depicts a conversational agent participating in an online support group with two human members A and B. Our support groups include 20 human members; they are purposefully small to promote trust, intimacy and accountability, but this means sometimes no one is available to respond to a post. In our example, A starts the conversation in the group chat, then after a few moments B responds back. Then A asks another question. Since no one replies to A within 3 minutes, the conversational agent generates a response. The conversational agent responds whenever someone explicitly directs a question to it or a message receives no response after a short period of time. Our online support group system facilitates dynamic interactions within group chats, allows discussions among multiple users, and ensures that relevant and timely support is provided in a collaborative environment. The focus of our research is on enhancing the accuracy of intent detection within these group chat settings, a crucial component for effective interactions.

A fully developed conversational agent typically consists of two primary components: intent classification (detection) and response generation. Intent classification involves identifying the user’s intent from their input, which is critical for understanding the context of the conversation (Whittaker et al., 2022; Rapp et al., 2021). Response generation, on the other hand, involves crafting a meaningful and contextually appropriate reply to the user’s input. However, the current version of our system focuses exclusively on intent classification. While this is a crucial part of the interaction, it does not yet include the capability to generate responses; but that will be our next step.

This paper aims to improve the intent classification component of a conversational agent. The methods and enhancements we introduce can be applied and replicated across various conversational agent systems. For this study, we use conversations among smoking-cessation support group members, primarily about using nicotine replacement therapy (NRT) to help them quit, as a case study to demonstrate the effectiveness of our approach in accurately detecting user intents in a specific context (Bendotti et al., 2023).

Recent advancements in NLP and LLMs, such as Llama 2, have opened new avenues for creating more sophisticated and responsive conversational agents. These models possess the ability to analyze and understand the nuanced emotions and intents expressed in user posts, enabling them to deliver interventions that are both empathetic and factual (Woebot, 2020; Brown, 2023). For example, when a user expresses feelings of failure or frustration with their quit attempt, the conversational agent can respond with encouragement and practical evidence-based tips, helping to mitigate the negative impacts of setbacks and stigma on the user’s journey to quit smoking (Kim et al., 2018).

The integration of NLP-based conversational agents into smoking cessation support groups represents a significant innovation in digital health. By utilizing advanced LLMs, these agents can bridge the gap between the need for personalized group-based support and the challenges posed by stigma. Moreover, they can enhance the overall quality of discussions within the group by providing accurate, evidence-based information, empowering users to take informed steps toward quitting (Lozano et al., 2020 and hashemitaheri et al.,2025).

This study presents an innovative approach to developing an NLP-based conversational agent designed to assist smokers in quitting. We address the challenge of insufficient high-quality data by employing a two-level data augmentation strategy: synthetic data augmentation and real data augmentation (Bayer et al., 2021; Khan et al., 2024). Having classified all 82,000 posts from our 45 prior smoking cessation support groups, we were left with numerous intents with low F1 (precision+recall) scores, below 80 percent. For these intents, we generate additional synthetic posts using prompt engineering with OpenAI’s ChatGPT model (Guo et al., 2024). Additionally, we scrape more than 10,000 posts from the Ex-Community, one of the main U.S. online communities for quitting smoking, to augment the dataset with real user-generated posts (Truth Initiative, 2024).

Performance evaluation of the retrained model demonstrates significant improvements in key metrics, confirming the effectiveness of our data augmentation approach (Dahmen et al., 2019). The main contribution of this paper is to describe an effective data augmentation approach to support the development of a smart conversational agent that will more accurately detect the intents of posts in support groups and related applications. By employing a novel two-level data augmentation approach, which integrates both synthetic and real data, we significantly improve the accuracy and effectiveness of intent detection by the conversational agent. This study not only provides a replicable framework for enhancing conversational agent performance in domains where data scarcity is a critical issue, but also contributes to the broader effort to create more engaging and stigma-free support groups for individuals seeking aid, e.g., with quitting smoking.

The rest of the paper is organized as follows: Section II reviews related studies on the use of conversational agents to increase engagement and lower stigma for smoking cessation, and the use of NLP-based conversational agents in digital health interventions generally. In Section III, we detail the methods used to enhance the accuracy of the smart conversational agent, focusing on the two-level data augmentation strategy involving both GPT-generated synthetic posts and real posts from a related context. Section IV presents the experimental setup, including the data collection, model training, and validation processes, followed by a discussion of the results. Finally, Section V offers concluding remarks and outlines potential future directions for research.

\section{Related Work}

\begin{figure*}[htbp]
  \centering
  \safeincludegraphics[width=0.8\textwidth]{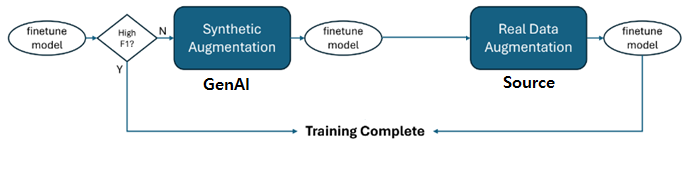}
  \caption{Overview of the proposed method}
  \label{fig:over}
\end{figure*}

Smoking cessation efforts have consistently faced challenges due to the stigma associated with smoking, which often manifests as negative societal judgments and stereotypes (Kim et al., 2018). This stigma not only impacts smokers' self-esteem but also diminishes their willingness to seek help and participate in cessation programs. Research has demonstrated that stigma exacerbates feelings of shame, guilt, and isolation, deterring individuals from engaging in cessation efforts and support groups. Research has demonstrated the potential benefits of online support groups for smoking cessation. A study by Graham et al. found that active participation in an online smoking cessation community was associated with higher quit rates, highlighting the importance of engagement in these support groups (Graham et al., 2017).

\begin{figure*}[htbp]
  \centering
  \safeincludegraphics[width=0.8\textwidth]{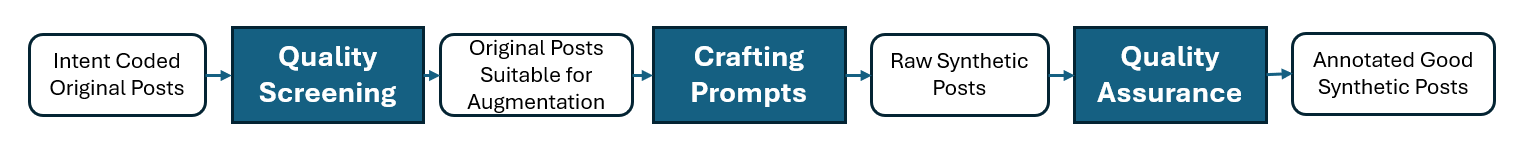}
  \caption{Overview of Synthetic Augmentation}
  \label{fig:synthetic-augmentation}
\end{figure*}

The use of online support groups for smoking cessation has gained popularity due to their accessibility and cost-effectiveness. However, these platforms often struggle with low user engagement and stigma-related issues. To address these challenges, researchers have been exploring the integration of conversational agents or chatbots into smoking cessation support systems.

\begin{figure*}[htbp]
  \centering
  \safeincludegraphics[width=0.8\textwidth]{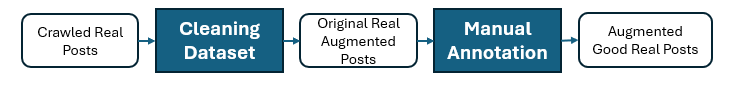}
  \caption{Overview of Real Augmentation}
  \label{fig:real-augmentation}
\end{figure*}

Conversational agents have emerged as a promising tool for motivating smoking cessation and providing continuous support to individuals attempting to quit. These AI-powered systems offer several advantages: enhanced accessibility and anonymity, 24/7 availability for support, and potential for personalized interactions. Recent studies have demonstrated the effectiveness of conversational agents in smoking cessation interventions. For instance, a pragmatic, multi-center, randomized clinical trial found that a treatment including a conversational agent was more effective than usual clinical practice in primary care for helping with tobacco cessation (Olano-Espinosa et al., 2022). The chatbot-assisted intervention showed promising results, although the outcomes were at the limit of statistical significance.

Motivational Interviewing (MI) has been identified as a particularly effective approach for conversational agent-based smoking cessation interventions. A study comparing MI-style and confrontational counseling-style conversational agent found that participants had a significantly higher overall rating for the MI conversational agent (He et al., 2024). The MI approach was especially effective in improving user experience-related outcomes, such as engagement, therapeutic alliance, and perceived empathy. Another study focused on developing an MI conversational agent with generative reflections to increase readiness to quit smoking. This research highlighted the potential of using advanced natural language processing techniques to create more personalized and empathetic responses, moving beyond scripted interactions (Brown et al., 2023).

Despite the promising results, developing effective conversational agent for smoking cessation faces several challenges such as: intent classification (accurately interpreting user utterances to provide appropriate responses), data scarcity (limited availability of high-quality labeled data for training models), and engagement (ensuring sustained user interaction with the conversational agent).

Giyahchi et al. (2023) addressed the challenge of low engagement in online health support groups by developing a model to detect individual intents while considering the overall context of group discussions. Their approach involved fine-tuning a BERT language model to account for the specific relationships between different intents in a group chat setting. This method aimed to enable conversational agents to contribute meaningfully to conversations, supporting health goals without disrupting the flow of dialogue. While this work represents a significant step forward in improving intent detection for health-related conversational agents, it is important to note that challenges in accuracy and adaptability to diverse user needs still persist in the field of conversational agents for healthcare applications (Giyahchi et al., 2023).

A notable advancement in the field of conversational agent-assisted smoking cessation is the development of QuitBot, as described by Bricker et al. (2024). This comprehensive conversational agent was created through an extensive 11-step user-centered design process, resulting in a 42-day structured quit-support program. QuitBot utilizes a vast library of over 11,000 pre-programmed question-and-answer pairs, supplemented by a GPT-3.5 model for handling more complex, open-ended queries. In a pilot randomized controlled trial, QuitBot demonstrated promising outcomes, with a 63\% 30-day point prevalence abstinence rate among participants who completed the program, compared to 38.5\% in the control group using the National Cancer Institute's SmokefreeTXT service (Bricker et al., 2024).

While these results are encouraging, the study highlights the ongoing challenges in conversational agent development, such as platform compatibility issues and limitations in handling open-ended questions. Despite its innovative approach and high user engagement, QuitBot's reliance on pre-programmed responses for many interactions suggests that there is still room for improvement in terms of accuracy and flexibility in addressing users' diverse needs. These persistent challenges in developing effective conversational agents for smoking cessation, including intent classification, data scarcity, and sustained user engagement, underscore the need for further research and innovation in this field.

To address these issues, we propose novel strategies that build upon existing work while introducing new techniques to enhance conversational agent performance. Our approach, which includes fine-tuning pretrained large language models, implementing advanced synthetic and real data augmentation techniques, and incorporating rigorous human validation processes, will be discussed in detail in the next section.

\section{Proposed Method}

Despite advances in NLP and machine learning, existing conversational agents for smoking cessation often struggle to accurately interpret user intents, especially within the dynamic and unstructured environment of group chats. This limitation is primarily due to insufficient high-quality annotated data, data noise, and the diverse ways individuals express themselves when discussing sensitive topics like smoking cessation. Consequently, these agents often fail to provide timely and contextually appropriate support, leading to decreased user engagement and reduced effectiveness of the intervention.

To address these challenges, we propose a comprehensive two-level data augmentation method aimed at enhancing the performance of conversational agents for support groups and related applications. We propose that our data augmentation method be used when the original finetuned model for the conversational agent has insufficiently low performance (F1) for detecting certain intents (topics). We recommend using Synthetic Augmentation (synthetic post generation using Generative AI) and, if the performance (F1) is still insufficient, using Real Data Augmentation (real posts from a different context, e.g., support group type), as shown in Figure~\ref{fig:over}. In our work, we use OpenAI's GPT-4 to create the synthetic post data and a national online support group to obtain additional real data for augmentation. The following sections detail each phase of the proposed method.

\subsection{Phase One: Synthetic Augmentation}

As illustrated in Figure~\ref{fig:over}, the initial dataset, referred to as Intent Coded Original Posts, consists of 82,000 posts obtained from our 45 prior online smoking cessation support groups. These posts were used in our prior project to develop our original intent detection model (Giyahchi et al., 2023). However, in our initial evaluations, we found that the model's performance was suboptimal for certain intents, likely at least in part due to the limited amount of available data. For the current work, we set a cutoff performance threshold of 80 percent for intent detection with the original posts, meaning that if the F1 score for any intent is below 80 percent, we view that as too low, necessitating data augmentation to improve intent detection.

For this data augmentation, we begin by generating synthetic posts from our current pool of Intent Coded Original Posts, ensuring that these new synthetic posts are good examples of the focal intents that require data augmentation (with low F1s). This synthetic data augmentation process consists of four stages: Quality Screening, Crafting Prompts, Generating Responses, and Quality Assurance.

\begin{table*}[htb]
\caption{Comparing performance metrics}
\centering
\resizebox{\textwidth}{!}{
\begin{tabular}{|l|cccc|cccc|cccc|c|c|c|}
 \hline
\multirow{2}{*}{\textbf{Intent Label}} & \multicolumn{4}{c|}{\textbf{Precision}} & \multicolumn{4}{c|}{\textbf{Recall}} & \multicolumn{4}{c|}{\textbf{F1}} & \multirow{2}{*}{\textbf{Orig}} & \multirow{2}{*}{\textbf{AugR}} & \multirow{2}{*}{\textbf{AugS}} \\
 & \textbf{Orig} & \textbf{Real} & \textbf{Synth} & \textbf{All} & \textbf{Orig} & \textbf{Real} & \textbf{Synth} & \textbf{All} & \textbf{Orig} & \textbf{Real} & \textbf{Synth} & \textbf{All} & & & \\ \hline \hline
nrt\_dontwork & 64.2 & 78.7 & 75.6 & 79.2 & 62.4 & 76.1 & 79.7 & 81.3 & 63.3 & 77.4 & 77.6 & 80.3 & 71 & 119 & 622 \\ \hline
nrt\_dreams & 71.4 & 81.0 & 85.7 & 85.7 & 78.1 & 85.0 & 90.0 & 90.0 & 75.0 & 82.9 & 87.8 & 87.8 & 50 & 27 & 388 \\ \hline
nrt\_howtouse & 66.7 & 85.2 & 81.5 & 84.6 & 58.7 & 74.2 & 71.0 & 68.8 & 62.7 & 79.3 & 75.9 & 75.9 & 158 & 120 & 594 \\ \hline
nrt\_itworks & 62.3 & 80.9 & 81.2 & 83.7 & 75.4 & 79.2 & 83.0 & 87.2 & 68.3 & 80.0 & 82.1 & 85.4 & 165 & 128 & 989 \\ \hline
nrt\_mouthIrr. & 72.7 & 72.7 & 81.8 & 90.0 & 71.6 & 80.0 & 90.0 & 81.8 & 71.6 & 76.2 & 85.7 & 85.7 & 39 & 14 & 101 \\ \hline
nrt\_nauseous & 77.8 & 77.8 & 77.8 & 77.8 & 83.1 & 77.8 & 87.5 & 77.8 & 80.2 & 77.8 & 82.4 & 77.8 & - & - & - \\ \hline
nrt\_od & 50.0 & 75.0 & 100.0 & 75.0 & 81.4 & 75.0 & 100.0 & 75.0 & 60.4 & 75.0 & 100.0 & 75.0 & 53 & 26 & 362 \\ \hline
nrt\_skinirr. & 75.0 & 76.9 & 84.6 & 84.6 & 79.4 & 90.9 & 91.7 & 91.7 & 75.7 & 83.3 & 88.0 & 88.0 & 41 & 17 & 70 \\ \hline
nrt\_stickissue & 93.3 & 87.5 & 93.3 & 93.3 & 80.3 & 82.4 & 77.8 & 82.4 & 86.8 & 84.8 & 84.8 & 87.5 & - & - & - \\ \hline
quitdate & 88.7 & 88.7 & 86.7 & 88.5 & 82.9 & 81.1 & 81.0 & 81.0 & 85.6 & 84.7 & 83.7 & 84.6 & - & - & - \\ \hline
ecigs & 84.2 & 83.3 & 77.8 & 83.3 & 96.7 & 83.3 & 77.8 & 83.3 & 88.8 & 83.3 & 77.8 & 83.3 & - & - & - \\ \hline
fail & 80.4 & 91.1 & 93.1 & 94.4 & 72.7 & 81.0 & 88.5 & 79.7 & 81.9 & 85.7 & 90.8 & 86.4 & - & - & - \\ \hline
scared & 90.9 & 90.9 & 90.9 & 90.9 & 75.9 & 80.0 & 80.0 & 80.0 & 83.1 & 85.1 & 86.7 & 85.1 & - & - & - \\ \hline
stress & 94.7 & 92.1 & 94.7 & 94.7 & 83.7 & 81.4 & 81.8 & 81.8 & 88.8 & 86.4 & 87.8 & 87.8 & - & - & - \\ \hline
tiredness & 69.2 & 83.3 & 84.6 & 83.3 & 76.4 & 83.3 & 91.7 & 83.3 & 72.2 & 83.3 & 88.0 & 83.3 & 119 & 93 & 77 \\ \hline
smokefree & 82.8 & 93.3 & 80.3 & 82.7 & 86.1 & 89.2 & 82.6 & 81.4 & 84.4 & 91.2 & 81.5 & 82.0 & - & - & - \\ \hline
smokingless & 94.7 & 90.0 & 89.5 & 90.0 & 82.4 & 90.0 & 81.0 & 85.7 & 88.1 & 90.0 & 85.0 & 87.8 & - & - & - \\ \hline
support & 91.7 & 88.2 & 90.2 & 90.4 & 86.7 & 88.8 & 86.0 & 86.8 & 89.1 & 88.5 & 88.0 & 88.6 & - & - & - \\ \hline
cigsmell & 88.5 & 77.8 & 84.0 & 92.0 & 93.7 & 91.3 & 80.8 & 92.0 & 81.0 & 84.0 & 82.4 & 92.0 & - & - & - \\ \hline
cravings & 68.1 & 76.3 & 81.1 & 83.0 & 71.9 & 88.2 & 83.3 & 86.1 & 69.8 & 81.8 & 82.2 & 84.5 & 269 & 874 & 57 \\ \hline
costs & 69.2 & 81.1 & 82.9 & 83.3 & 86.7 & 90.9 & 82.9 & 88.2 & 76.7 & 85.7 & 82.9 & 85.7 & 116 & 137 & 85 \\ \hline
health & 68.4 & 78.1 & 81.7 & 81.7 & 88.2 & 83.8 & 81.7 & 84.1 & 77.1 & 80.9 & 81.4 & 82.9 & 267 & 610 & 588 \\ \hline
weightgain & 67.3 & 82.2 & 79.1 & 84.1 & 84.6 & 83.6 & 75.6 & 86.0 & 74.8 & 83.1 & 77.3 & 85.1 & 118 & 95 & 146 \\ \hline
\textbf{AVE.} & \textbf{77.1} & \textbf{83.1} & \textbf{84.8} & \textbf{86.1} & \textbf{79.9} & \textbf{83.3} & \textbf{83.4} & \textbf{83.8} & \textbf{78.0} & \textbf{83.1} & \textbf{83.9} & \textbf{84.8} & - & - & - \\ \hline
\end{tabular}
}
\label{tab:comp}
\end{table*}

\begin{figure*}[htb]
  \centering
  \safeincludegraphics[width=0.8\textwidth]{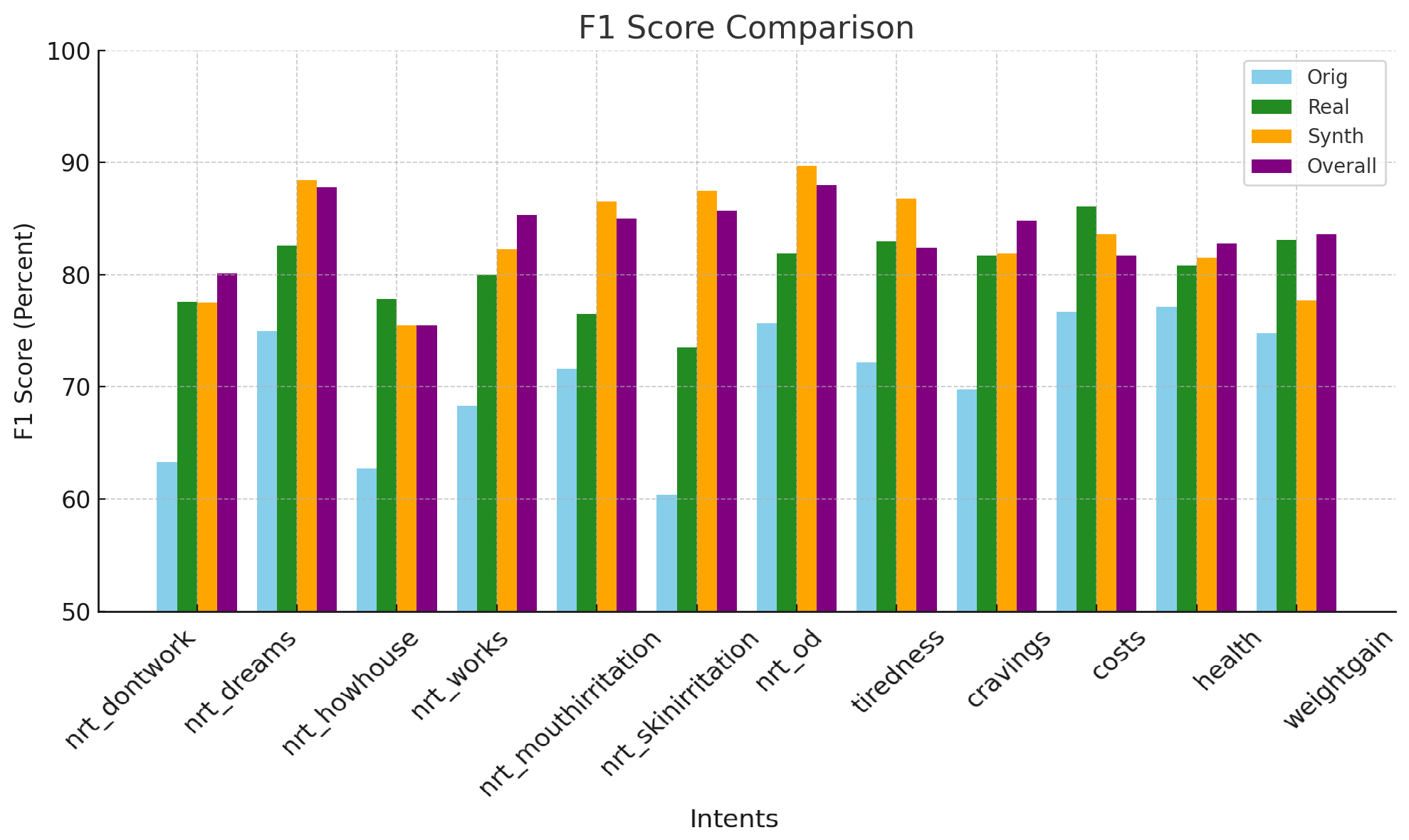}
  \caption{Comparing the F1 Score of Classification for Low-accuracy Intents}
  \label{fig:f1}
\end{figure*}

\begin{table*}[htbp]
\centering
\caption{Data augmentation summary}
\begin{tabular}{|l|c|c|c|c|c|c|} 
\hline
Intent Label & Orig Posts & Screened & Raw Synth & Good Synth & Orig Real & Good Real \\ \hline \hline
nrt\_dontwork & 481 & 361 & 689 & 622 & 101 & 89 \\ \hline
nrt\_dreams & 398 & 329 & 424 & 388 & 31 & 27 \\ \hline
nrt\_howtouse & 577 & 424 & 801 & 594 & 134 & 120 \\ \hline
nrt\_itworks & 955 & 764 & 1000 & 983 & 137 & 128 \\ \hline
nrt\_mouthirritation & 207 & 146 & 101 & 98 & 22 & 14 \\ \hline
nrt\_od & 69 & 62 & 355 & 309 & 29 & 26 \\ \hline
nrt\_skinirritation & 241 & 114 & 99 & 70 & 27 & 24 \\ \hline
tiredness & 247 & 137 & 248 & 77 & 99 & 93 \\ \hline
cravings & 2198 & 479 & 221 & 57 & 985 & 845 \\ \hline
costs & 697 & 128 & 201 & 136 & 104 & 93 \\ \hline
health & 1400 & 308 & 401 & 378 & 708 & 581 \\ \hline
weightgain & 875 & 173 & 200 & 145 & 110 & 95 \\ \hline
\textbf{AVERAGE} & \textbf{691} & \textbf{301} & \textbf{424} & \textbf{370} & \textbf{208} & \textbf{153} \\ \hline
\end{tabular}
\label{tab:data}
\end{table*}

\begin{figure*}[htbp]
  \centering
  \safeincludegraphics[width=0.8\textwidth]{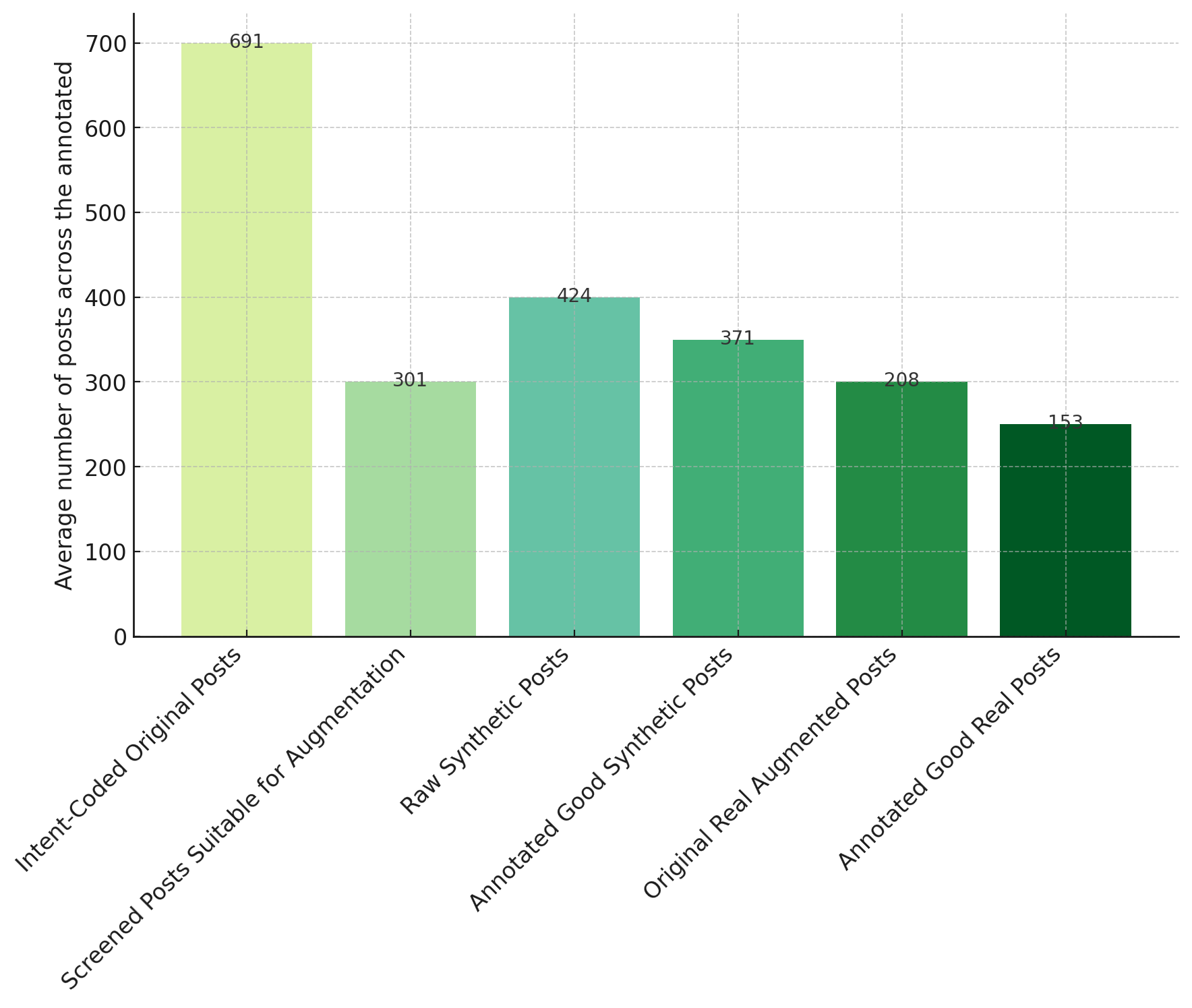}
  \caption{Average Number of Posts Across the Annotated Labels}
  \label{fig:avg}
\end{figure*}

\subsubsection{Quality Screen} Due to the variability inherent in human posts in support groups, not all posts in the original dataset are suitable for creating synthetic posts. Therefore, inspired by previous works, we implement a quality screening step aimed at isolating the most relevant and contextually rich posts (Wei et al., 2019). Each post is evaluated manually by an expert to ensure it aligns with a focal smoking cessation intent and has the potential to contribute meaningfully to synthetic data generation.
\begin{itemize} \item \textbf{Relevance:} Each post is assessed by a human expert for its alignment with a predefined smoking cessation intent or topic. Posts must either explicitly or implicitly address the focal topic like craving management, withdrawal symptoms, or nicotine replacement therapy. For example, a post like `Feeling stressed, but trying not to smoke'' aligns with the craving management intent, while a post like I love this new nicotine gum flavor'' is excluded for focusing on product preference rather than the cessation process. Posts must contribute meaningfully to the smoking cessation conversation by offering personal insights or actionable advice. For instance, `Quitting is hard, but keeping busy helps me manage cravings'' is considered relevant, while generic posts like Quitting is hard'' are excluded unless they provide context or advice. \item \textbf{Contextual Completeness:} Each post is further evaluated to ensure it conveys a complete and coherent thought and can stand alone without requiring prior or subsequent posts for clarity. Posts missing key elements—such as the challenge, action, or outcome—are flagged as incomplete. For example, a post like `Cravings are tough'' is discarded unless it provides additional context or coping mechanisms, as in Cravings are tough, but deep breaths help.'' \item \textbf{Clarity:} Posts are additionally evaluated to ensure they are clear, concise, and easy to understand without unnecessary noise such as hashtags, links, or jargon. Posts with extraneous elements that obscure the intended message are excluded. The end result of the quality screening process is what we call Prescreened Posts for Augmentation. \end{itemize} \subsubsection{Crafting Prompts} After posts pass through quality screening, a human expert crafts prompts to guide the GPT model in generating similar synthetic posts. The key to prompt crafting is to reflect the core idea of each post while allowing the GPT model to generate varied and contextually relevant posts. For instance, a post like `When cravings hit, I go for a walk to distract myself'' is transformed into What are some effective ways to manage cravings when trying to quit smoking like walking?'' To ensure diversity, prompts are designed to encourage a range of generated responses, avoiding repetition. \subsubsection{Generating Responses} Once prompts are finalized, they are input into the GPT model to generate multiple responses per prompt. The goal is to create synthetic posts that accurately mimic the topic, style, and tone of the original posts. For example, from the prompt `What are some effective ways to manage cravings when trying to quit smoking like walking?'', the model generates responses like Chewing gum helps distract me from cravings'' and Going for a walk clears my mind when the urge to smoke hits.'' The expert reviews initial outputs, refines prompts as necessary, and filters out responses that do not meet the criteria. The resultant dataset is what we call Raw Synthetic Posts. \subsubsection{Quality Assurance} Each synthetic post is independently evaluated by two human annotators trained to assess whether it fits the focal intent definition, is logical and fluent, and not overly repetitive. If annotators disagree, they discuss and attempt to reach agreement; if they cannot, the post is escalated to an expert judge. Through this rigorous quality assurance process, we ensure that only synthetic posts that are on-topic, clearly worded, logical, and nonrepetitive are included. These usable synthetic posts are what we call Annotated Good Quality Synthetic Posts. \subsubsection{Stopping Criteria for Synthetic Post Generation} We establish clear stopping criteria based on two factors: semantic drift and redundancy. \begin{itemize} \item \textbf{Semantic Drift:} Semantic drift occurs when generated posts deviate from intended topics, leading to generic or off-topic responses. When detected, the expert refines prompts to refocus generation. \item \textbf{Redundancy:} Redundancy occurs when responses are too similar in content, reducing diversity. When redundancy becomes significant, prompts are adjusted to encourage more varied content. \end{itemize} \subsection{Phase Two: Real Data Augmentation} After augmenting the dataset with synthetic posts, we incorporate real-world data to further diversify and enrich the dataset. We scrape over 10,000 posts from the Ex-Community (Truth Initiative, July 2024). This real-world data undergoes steps to ensure that it meets the quality standards necessary for effective model training. As shown in Figure~\ref{fig:real-augmentation}, the two critical steps are Cleaning the Dataset and Manual Annotation. \begin{itemize} \item \textbf{Cleaning Dataset:} Irrelevant and noisy content is removed (incomplete posts, non-English messages, spam, off-topic content). Text is normalized (spelling/punctuation/formatting) so posts are comparable to the original and synthetic data. \item \textbf{Manual Annotation:} Two human annotators independently label each relevant part of each post as fitting one of the original smoking-related intents or not fitting any focal intent. Disagreements are resolved via discussion or an expert judge. Posts annotated as fitting focal intents are included as Augmented Good Real Posts. \end{itemize}

\section{Experimental Evaluation} In this section, we present results evaluating the effectiveness of the two-level augmentation strategy on intent detection performance. \subsection{Evaluation Metrics} \begin{figure*}[htbp] \centering \includegraphics[width=\textwidth]{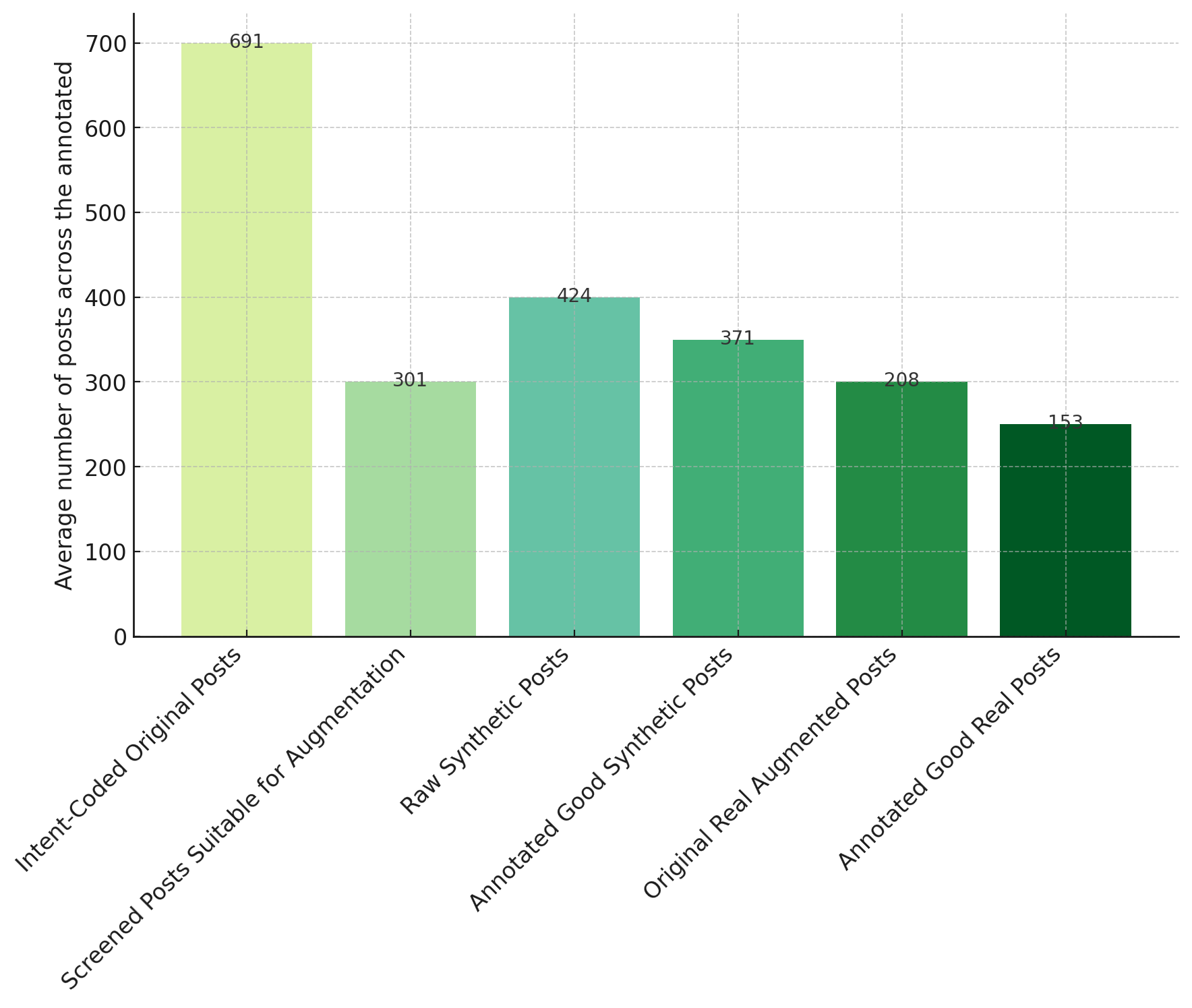} \caption{Average Number of Posts Across the Annotated Labels} \label{fig:avg} \end{figure*} The experiments are conducted using the augmented dataset, which is created by integrating both synthetic and real-world data, as described in the previous section. The results demonstrate the extent to which the data augmentation strategy improves the conversational agent’s performance compared to baseline models that were trained on the original, unaugmented dataset. We assess the model’s performance across various standard NLP metrics, namely its \emph{precision}, \emph{recall}, and \emph{F1 scores}, to see if it can accurately detect user intents within the challenging environment of group chats. \begin{description} \item[Precision:] Precision is a measure of the accuracy of the positive predictions made by the model. It is calculated as the ratio of correctly predicted positive observations compared to the total predicted positive observations. \[ \text{Precision} = \frac{TP}{TP + FP} \tag{1} \] \item[Recall:] Recall, also known as sensitivity or the true positive rate, measures the model's ability to correctly identify all relevant instances. It is calculated as the ratio of correctly predicted positive observations compared to all observations in the actual class. \[ \text{Recall} = \frac{TP}{TP + FN} \tag{2} \] \item[F1 Score:] The F1 score is the harmonic mean of precision and recall. \[ F1 = \frac{2 \cdot (\text{Precision} \cdot \text{Recall})}{\text{Precision} + \text{Recall}} \tag{3} \] \end{description} \subsection{Experimental Results}

Table~\ref{tab:comp} lists the 23 smoking-related intents we trained the original model to detect, and the 12 intents that require data augmentation due to low F1 values (\(<80\)). Table~\ref{tab:comp} shows precision, recall, and F1 for each intent label. Figure~\ref{fig:f1} illustrates the results presented in Table~\ref{tab:comp}.
Figure~\ref{fig:avg} displays the average number of posts for original posts, screened posts for augmentation, raw synthetic posts, annotated good synthetic posts, original real augmented posts, and annotated good real augmented posts.
\section*{Acknowledgements}
We would like to thank the student researchers Amber Jiang, Aniket Pratap, Farrah Galal, Madison Reed, Brandon Cruz, and Ronin Vohra who helped us annotate both the synthetic and real data, enabling us to reach the desired level of performance.

We would like to thank the EX Community for providing open-source smoking cessation forum data. Their platform was invaluable for this research.

Lastly, we would like to acknowledge the use of ChatGPT for proofreading this paper.

\end{document}